# A Comparative Study of Sampling Methods with Cross-Validation in the FedHome Framework

Arash Ahmadi[1], Sarah S. Sharif[1], and Yaser M. Banad[1*]
School of Electrical and Computer Engineering, University of Oklahoma, Norman, OK, 73019
*Corresponding author: bana@ou.edu

*Abstract*— **This paper presents a comparative study of sampling methods within the FedHome framework, designed for personalized in-home health monitoring. FedHome leverages federated learning (FL) and generative convolutional autoencoders (GCAE) to train models on decentralized edge devices while prioritizing data privacy. A notable challenge in this domain is the class imbalance in health data, where critical events such as falls are underrepresented, adversely affecting model performance. To address this, the research evaluates six oversampling techniques using Stratified K-fold cross-validation: SMOTE, Borderline-SMOTE, Random OverSampler, SMOTE-Tomek, SVM-SMOTE, and SMOTE-ENN. These methods are tested on FedHome's public implementation over 200 training rounds with and without stratified K-fold cross-validation. The findings indicate that SMOTE-ENN achieves the most consistent test accuracy, with a standard deviation range of 0.0167-0.0176, demonstrating stable performance compared to other samplers. In contrast, SMOTE and SVM-SMOTE exhibit higher variability in performance, as reflected by their wider standard deviation ranges of 0.0157-0.0180 and 0.0155-0.0180, respectively. Similarly, the Random OverSampler method shows a significant deviation range of 0.0155-0.0176. SMOTE-Tomek, with a deviation range of 0.0160-0.0175, also shows greater stability but not as much as SMOTE-ENN. This finding highlights the potential of SMOTE-ENN to enhance the reliability and accuracy of personalized health monitoring systems within the FedHome framework.**

## I. INTRODUCTION

Federated learning (FL) has emerged in recent years as a distributed machine learning approach, enabling collaborative model training while safeguarding user data privacy [1]. In FL, a shared global model is collaboratively trained across numerous decentralized edge devices or servers, each maintaining local data. These devices transmit model updates to a central server, which integrates these updates into the global model. This innovative approach allows users to jointly reap the benefits of a collective model while preventing any raw user data from being revealed. FL thus provides an elegant solution for training machine learning models in a privacy-centric manner, particularly beneficial for data-sensitive sectors like healthcare, finance, and mobile technology [2].

One notable application of FL is in the domain of in-home health monitoring, aiming to offer continuous, personalized care within the comfort of users' homes. Figure 1 shows the high-level implementation of federated learning in this context. Given the sensitive nature of health data, users often hesitate to share it with external entities. To address this challenge, a novel framework called FedHome was proposed to leverage federated learning for personalized in-home health monitoring [3]. FedHome adopts a cloud-edge architecture, where the edge comprises user homes equipped with data-collecting devices like wearables and smartphones. Instead of transmitting sensitive health data, FedHome leverages FL to train a comprehensive global model on a central cloud server using data from various homes. This strategy effectively utilizes extensive data for model accuracy while keeping sensitive health data localized at the edges to preserve privacy. This framework could pave the way for broader applications in other fields where data sensitivity is paramount, illustrating the versatile potential of federated learning.

FedHome uses a Generative Convolutional Autoencoder (GCAE) as its primary machine learning model, trained with the FL approach. An autoencoder is a kind of neural network whose main purpose is to compress the input into a lower-dimensional and informative representation and then restore it to a similar output as the original input [4]. GCAE integrates a convolutional pyramid encoder, a convolutional decoder, and a Multi-Layer Perceptron (MLP) to learn low-dimensional, common, and representative features of high-dimensional healthcare data. The encoder-decoder structure enables the model to reconstruct the input data, while the MLP predicts the class label [3]. Upon developing the global model on the cloud, FedHome personalizes it for each home by retraining it with localized user data. This process ensures the model captures unique characteristics, leading to precise, customized in-home monitoring, such as detecting falls in elderly individuals. The overarching goal is to perform privacy-preserving yet personalized health monitoring using federated learning.

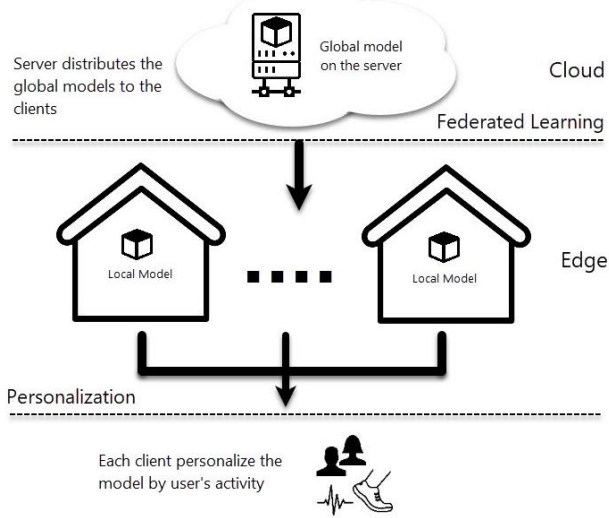

**Fig. 1.** FedHome framework architecture diagram showing cloud server, edge devices, and how global and personalized models are trained.

A key challenge in medical datasets is that user data often exhibits imbalanced class distribution [6]. Critical instances like abnormal states are underrepresented compared to more frequent routine activities. For example, elderly falls during in-home monitoring are typically sparse compared to routine activities like walking or sleeping.

This class imbalance can degrade model performance as machine learning algorithms are biased towards majority classes [7]. Consequently, the model might overlook or inadequately learn about crucial yet less frequent events like falls. To fix this issue, sampling techniques are employed to balance the distribution between the minority and majority classes, enhancing the model's ability to recognize and respond to critical events effectively. Integrating these balancing techniques with GCAE's architecture reveals the potential of FedHome to significantly improve the safety and well-being of individuals, especially in sensitive applications like elderly care.

Sampling techniques can be broadly classified into oversampling and undersampling [8]. Oversampling methods increase the number of minority class samples by replicating existing ones or generating new synthetic ones. Undersampling methods reduce the number of majority class samples by randomly removing them or applying some criteria to select the most representative ones. Both oversampling and undersampling aim to balance the class distribution and improve the performance of classifiers on imbalanced datasets. Several oversampling techniques can be applied:

- **SMOTE (Synthetic Minority Oversampling Technique):** It utilizes the k-nearest neighbors (kNN) approach to generate new, artificial samples of the minority class [9]. By identifying k nearest neighbors within the existing minority data points, SMOTE "interpolates" new samples along the lines connecting these neighbors, effectively increasing the minority class size without simply replicating existing data.
- **Borderline-SMOTE:** Focuses on oversampling minority samples near the decision boundary between classes, which are more prone to misclassification [10]. It uses an SVM algorithm to detect the borderline samples and generates synthetic samples along their connecting line segments.
- **Random Oversampler:** Increases the proportion of minority class samples by replicating them randomly. Although simple and initially effective, it can lead to overfitting and reduce class diversity. Widely utilized in research [11] and included in the imbalanced-learn Python library [12].
- **SVM-SMOTE:** Produces new minority samples along the hyperplanes between classes learned by an SVM model [13]. It can capture the complex distribution of the minority class without generating noisy samples in sparse regions.

Oversampling and undersampling techniques can also be combined, which is called hybrid methods [8]. Since the FedHome framework aims to reconstruct the model using the oversampling techniques, we also wanted to see how these hybrid methods can perform in this situation. Some of these methods are:

- **SMOTE-ENN:** This method combines oversampling with SMOTE and undersampling with Edited Nearest Neighbors (ENN). ENN removes both majority and minority samples whose class differs from the majority class of their nearest neighbors, resulting in a smoother decision boundary and potentially improved classification accuracy. Paper [14] used SMOTE-ENN for imbalanced medical data classification.
- **SMOTE-Tomek:** This hybrid approach combines oversampling with SMOTE and undersampling with Tomek Links. Tomek Links are pairs of neighboring samples from different classes. Removing these samples can reduce noise or outliers and improve class separability. The authors of [15] applied SMOTE-Tomek for diabetes data classification using the Random Forest method.

Several studies have explored the application of sampling techniques to address class imbalance in various domains. Two interesting papers, "Comparing Different Resampling Methods in Predicting Students' Performance Using Machine Learning Techniques" by Ghorbani and Ghousi [17] and "Oversampling Techniques for Diabetes Classification: a Comparative Study" by Mesquita et al. [16], provide valuable insights into the effectiveness of these methods.

Ghorbani and Ghousi [17] investigated the impact of class imbalance on machine learning models' performance in predicting students' academic outcomes. They compared six resampling techniques: SMOTE, Borderline-SMOTE, Random Over Sampler, SMOTE-ENN, SVM-SMOTE, and SMOTE-Tomek, across various classifiers such as Random Forest, K-Nearest Neighbor, and Artificial Neural Networks. The study employed different evaluation metrics to assess the performance of the models. The results showed that the classifiers' performance improved when using balanced datasets generated by the resampling methods. Furthermore, the authors found that the SVM-SMOTE technique outperformed other resampling methods based on the Friedman statistical test. The Random Forest classifier achieved the best results when combined with SVM-SMOTE.

Similarly, Mesquita et al. [16] conducted a comparative study of oversampling techniques for diabetes classification using the PIMA Indian Diabetes Data Set. They tested six oversampling methods: ADASYN, Borderline SMOTE, KMeans SMOTE, Random Over Sampler, SMOTE, and SVM SMOTE, in conjunction with ten machine learning algorithms. The study aimed to identify the most effective combination of oversampling technique and classifier for predicting diabetes. The results indicated that the combination of SVM SMOTE and AdaBoost algorithm achieved the highest accuracy, recall, precision, and specificity while maintaining a low false negative rate.

These studies exemplify the critical need to address class imbalance in datasets to enhance the performance of machine learning models. They demonstrate that oversampling techniques can effectively balance the class distribution, leading to better classification results. The comparative analysis of various resampling methods provides valuable guidance for selecting the most suitable technique based on the specific dataset and application domain.

Federated learning introduces its own set of challenges, such as Non-IID data distributions and communication constraints, potentially influencing the effectiveness of oversampling techniques. Non-IID is an acronym for non-independent and identically distributed data. This is a situation where the data points either have some dependence among them, or they follow different probability distributions, or both. A common example of non-IID data is when the data is collected from different users of a mobile device. Each user may have their own preferences and behaviors, which affect the data they generate. Therefore, the data from one user may not be representative of the data from another user, or the data from the whole population [5]. In health monitoring, class imbalance presents a unique problem because it's crucial to learn from the minority classes accurately. Therefore, a dedicated analysis is necessary to identify the most suitable oversampling approach for the FedHome system. This paper aims to provide such an analysis by examining prominent oversampling methods within the context of personalized health monitoring.

The goal of this study is to evaluate and compare key sampling techniques in the context of FedHome. This will help determine the most effective approach for dealing with imbalance in personalized in-home health monitoring data. The FedHome framework leverages federated learning and generative convolutional autoencoders to develop a global decentralized model based on data collected from edge devices in homes. This research will assess the performance of various oversampling and undersampling techniques using metrics such as using cross validation for accuracy and the standard deviation of test accuracy. By pinpointing the most effective data sampling strategies, we can improve the precision and reliability of health monitoring systems, ultimately leading to better health outcomes. Furthermore, the paper will discuss the advantages and disadvantages of each technique and suggest directions for future research in this domain. This study stands at the intersection of machine learning innovation and practical healthcare application, highlighting the transformative potential of distributed intelligence in personal health management.

## II. METHODS

*FedHome Overview*

This section provides a high-level overview of the FedHome framework and the Generative Convolutional Autoencoder (GCAE) model that it utilizes. Within this framework, a GCAE model is trained across these homes and the cloud using federated learning, ensuring that raw user data is not shared. The encoder component of the model is responsible for extracting low-dimensional features from the input data, which are usually high-dimensional and complex signals from sensors or medical images. The encoder network consists of several convolutional and max-pooling layers that reduce the dimensionality and capture the representative features of the data. Using the SMOTE oversampling technique, the decoder then uses these latent features to reconstruct the original input. The decoder network is a symmetric counterpart of the encoder network, consisting of several deconvolutional and up-sampling layers that increase the dimensionality and recover the original data. The reconstruction loss between the input and the output of the decoder is used to optimize the encoder and the decoder jointly. Finally, a Multilayer Perceptron (MLP) classifier is used to perform prediction using the features extracted by the encoder. The MLP network consists of several fully connected layers that map the latent features to the output classes. The prediction loss between the actual label and the predicted probability distribution is used to optimize the MLP network. The GCAE model is trained in an end-to-end manner by minimizing the weighted sum of the reconstruction loss and the prediction loss.

*Oversampling in FedHome*

The FedHome framework applies oversampling before the personalization stage, as shown in Figure 2. Specifically, once the cloud model has been trained, the minority class in each user's skewed local dataset is oversampled. The FedHome framework adopts a popular oversampling method called SMOTE (Synthetic Minority Over-sampling Technique), which creates new samples by interpolating between the existing samples of the minority class. The oversampling process is performed in the low-dimensional latent space of the GCAE encoder. This encoded representation of the data makes the oversampling process more effective, as it preserves the essential information and reduces the noise and redundancy of the data. As a result, the personalized model can learn effectively from the reconstructed class-balanced data, leading to improved performance and accuracy. This approach to addressing class imbalance has been shown to be highly effective in the FedHome framework [2].

## III. INCORPORATING STRATIFIED K-FOLD CROSS-VALIDATION IN FEDHOME FRAMEWORK

The inherent class imbalance in personalized health monitoring data poses a significant challenge, impacting the generalizability and performance of machine learning models developed within the FedHome framework. To address this, incorporating Stratified K-Fold cross-validation into the model validation process provides an additional layer of robustness against class imbalance. Figure 3 illustrates stratified K-Fold cross-validation, showing how the train and test sets (blue and orange) are split and where the rare data (green area) is located within the entire dataset. This approach ensures the inclusion of rare data in at least one fold, unlike the traditional train-test split where rare data might be excluded from one of the sets.

*Stratified K-Fold*

Stratified K-Fold cross-validation is an extension of the traditional K-Fold technique, specifically designed to maintain the original distribution of classes in each fold. This approach ensures that each fold is a good representative of the entire dataset, particularly important in the context of imbalanced datasets, as it preserves the proportion of the minority class, thereby maintaining model accuracy and reliability.

*Methodology*

In Stratified K-Fold cross-validation, the dataset is split into K different subsets (or folds). Each iteration uses one subset as the test set, while the remaining subsets are combined to form the training set. What distinguishes Stratified K-Fold from its non-stratified counterpart is the stratification process, where the splits are made by preserving the percentage of samples for each class, ensuring that each fold reflects the overall class distribution of the dataset.

This validation technique can be particularly beneficial for the FedHome framework for several reasons:

- **Model Evaluation:** It provides a more accurate measure of model performance and generalizability across different subsets of data, accounting for the class imbalance.
- **Bias Reduction:** By ensuring that each fold contains a representative distribution of classes, Stratified K-Fold helps reduce bias toward the majority class, allowing for more balanced learning and prediction.

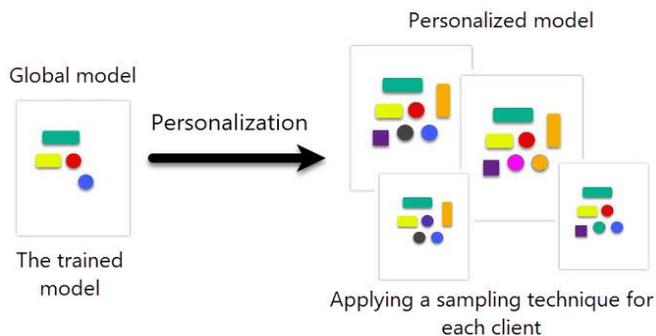

**Fig. 2.** Illustration of how oversampling and undersampling techniques modify and add samples of minority class.

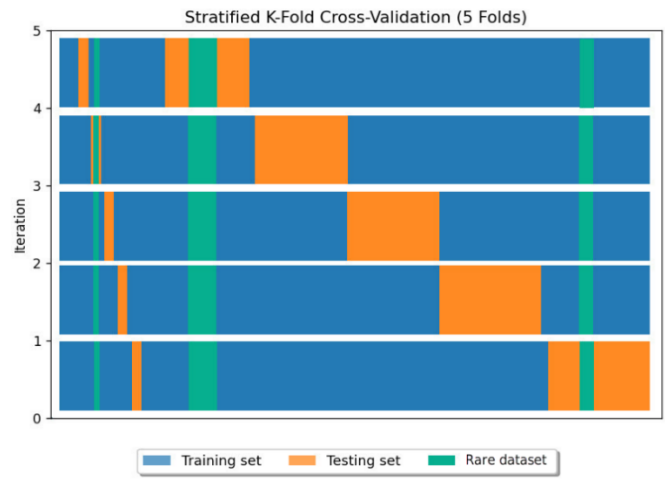

**Fig. 3.** Stratified K-Fold Cross Validation. It shows how the train and test set (blue and orange) is being split, and where the rare data (green area) is located in the whole dataset. This approach ensures the inclusion of the rare dataset at least in one-fold, compared to the traditional train-test split where the rare data could not be included in one of the sets.

*Techniques for saving the model*

Our implementation uses selected folds for testing and training datasets to evaluate each sampling technique. The global and personalized model would be then trained on their corresponding folds, and for testing different sampling techniques for that fold in the personalization phase, we can reload the non-personalized model after the whole training is done. This will save time when testing all folds for the different sampling techniques. The original implementation of the **save_global_model** and **load_model** methods in the serverbase.py file in the FedHome Github repository [18] (which is an extended version of this GittHub repository [19] by the same author) is inaccurate and does not follow best practices for saving and loading models in PyTorch, particularly in the context of cross-validation. The previous approach only saves and loads the **global_model** attribute of the server object, neglecting other crucial parameters and components required to reproduce the model's state accurately. These parameters include the list of clients, selected clients, slow clients for training and sending, and performance metrics such as test accuracy, test Area Under the Curve (AUC), and training loss.

In the updated implementation, the **save_global_model** method uses **torch.save** to save a dictionary containing the model's state dictionary (**global_model.state_dict**()) along with the aforementioned parameters. The **load_model** method, on the other hand, loads the saved checkpoint using **torch.load** and restores the model's state using the **load_state_dict** method. It also retrieves the saved parameters from the checkpoint dictionary and assigns them to their respective attributes in the server object.

This comprehensive saving and loading of the model and its associated parameters are especially critical for cross-validation. In cross-validation, the model needs to be evaluated on multiple folds of the data, requiring the model to be reset to its previous state before evaluating each fold. By saving and loading the complete model state, including all relevant parameters, the revised implementation ensures the model can be accurately restored to its previous state for each fold evaluation.

```
Algorithm 1 Updated methods for saving the state of the model
1:  procedure SAVE-GLOBAL-MODEL
2:     save global-model-state-dict, clients, selected-clients, train-slow-clients,
       send-slow-clients, rs-test-acc, rs-test-auc, rs-train-loss to file
3:  end procedure
4:  procedure LOAD-MODEL
5:     load global-model-state-dict, clients, selected-clients, train-slow-clients,
       send-slow-clients, rs-test-acc, rs-test-auc, rs-train-loss from file
6:     restore global model state and assign loaded parameters to attributes
7:  end procedure
8:  procedure SAVE-CLIENT-MODEL
9:     save model-state-dict, train-samples, test-samples, train-slow, send-slow,
       train-time-cost, send-time-cost to file
10: end procedure
11: procedure LOAD-CLIENT-MODEL
12:    load model-state-dict, train-samples, test-samples, train-slow, send-slow,
       train-time-cost, send-time-cost from file
13:    restore client model state and assign loaded parameters to attributes
14: end procedure
```

**Algorithm. 1.** Pseudo-code for updated save and load methods in serverbase.py and clienthome.py

Furthermore, considering the evaluation of the model depends on the clients in the federated learning setting, it is essential to implement the **save_client_model** and **load_client_model** methods for each client individually. These methods save and load the client-specific model state dictionary, training and testing samples, and slow client flags. The revised implementation of the **save_global_model**, **load_model**, **save_client_model,** and **load_client_model** methods demonstrate a more comprehensive and robust approach to saving and loading models in the federated learning framework, particularly for cross-validation purposes. It ensures that all necessary parameters and components are properly saved and restored, enabling reproducibility and accurate evaluation of the model's performance across different folds of the data.

*Integration with FedHome*

To integrate Stratified K-Fold cross-validation within the FedHome framework, the following steps are proposed:

- **Stratification:** Applying Stratified K-Fold cross-validation should be done before training the model to ensure the consistency between the global model and personalized model
- **Evaluation Metrics:** Utilize balanced evaluation metrics such as AUC, and Accuracy to provide a comprehensive view of model performance across all classes.

Adopting Stratified K-Fold cross-validation can further enhance its capability of FedHome framework to develop and validate accurate and generalizable models across diverse and imbalanced datasets encountered in personalized health monitoring scenarios.

## IV. IMPLEMENTATION DETAILS

*Used libraries*

The analysis and evaluation of the different sampling techniques were implemented by modifying the public implementation of the FedHome framework, available on GitHub [18]. The following classes from the imbalanced-learn library [20] were integrated into the FedHome codebase:

- **imblearn.over_sampling.SMOTE:** For applying the SMOTE oversampling technique.
- **imblearn.over_sampling.RandomOverSampler:** For oversampling the minority class by random replication.
- **imblearn.combine.SMOTEENN:** For applying the hybrid SMOTE-ENN technique.
- **imblearn.combine.SMOTETomek:** For applying the hybrid SMOTE-Tomek links technique
- **imblearn.over_sampling.SVMSMOTE**: For applying SMOTE with SVM-based synthetic sample generation.
- **imblearn.over_sampling.BorderlineSMOTE:** For applying SMOTE focusing on borderline instances of the minority class.

These classes were used to implement the various sampling methods within the FedHome framework's codebase. The oversampling was applied before the personalization stage, as described in Section II.

*Train function*

The original train function of the public implementation of FedHome [18] has been modified to incorporate cross-validation and support different sampling techniques, as shown in **Algorithm 2**. The updated train function begins by training the global model for a specified number of rounds (self.global_rounds) for a specific fold. The cross-validation process is carried out using a fixed number of folds (num_folds), typically set to 5. During each round, the selected clients perform local training on their respective data, and the server aggregates the updated parameters to refine the global model. This process continues until the global model converges or the maximum number of rounds is reached (which is 200 rounds that have been explored in this paper)

Once the global model training is complete, the function proceeds to perform cross-validation for each sampling technique specified in self.list_oversampling_method. For each sampling technique on a specific fold, the clients generate their data using the corresponding sampling method (client.generate_data). This step ensures the data is appropriately oversampled or undersampled based on the selected technique. The generate_data function considers the current fold (fold_now) and the total number of folds (num_folds).

After data generation, the selected clients are set to include all clients (self.selected_clients = self.clients), and the personalized model training begins. The training process is similar to the global model training, where the clients perform local training on their personalized data for a specified number of rounds. During each round, the evaluation metrics (accuracy, AUC, standard deviation of accuracy, and standard deviation of AUC) are calculated using the evaluate_folds function, which takes into account the current fold and the total number of folds.

The evaluation metrics for each sampling technique and each round are stored in separate arrays (self.accuracies, self.aucs, self.std_accuracies, and self.std_aucs). These arrays allow tracking of the personalized models' performance across different sampling techniques and folds.

After completing personalized model training for a specific sampling technique, the global model (un-personalized model) is reloaded (self.global_model = self.load_model()) instead of being discarded, as is the case in original cross-validation implementations to continue testing other sampling technique for the current fold. Additionally, the client models are reloaded (client.load_client_model()) to allow for another sampling technique. Once this algorithm is completed, it will be executed again for a different fold.

Finally, the average evaluation metrics for all folds are plotted using the matplotlib library. Separate plots are generated for accuracy, AUC, standard deviation of accuracy, and standard deviation of AUC. Each plot displays the performance curves for different sampling techniques, allowing for a visual comparison of their effectiveness. The updated implementation of the train function, along with the proper saving and loading of models, enables comprehensive cross-validation for multiple sampling techniques in the federated learning framework.

```
Algorithm 2 Cross-validation workflow for multiple sampling techniques
 1: procedure TRAIN
 2:     Train global model for self.global-rounds for a specific fold
 3:     Save global model and client models
 4:     for sampling-technique in self.list-oversampling-method do
 5:         for client in self.clients do
 6:             client.generate-data(sampling-technique, fold, num-folds)
 7:         end for
 8:         self.selected-clients ← self.clients
 9:         for round in range(self.global-rounds) do
10:             if round mod self.eval-gap == 0 then
11:                 accuracy, auc, std-accuracy, std-auc ← self.evaluate-folds(fold, num-folds)
12:                 self.accuracies[sampling-technique].append(accuracy)
13:                 self.aucs[sampling-technique].append(auc)
14:                 self.std-accuracies[sampling-technique].append(std-accuracy)
15:                 self.std-aucs[sampling-technique].append(std-auc)
16:             end if
17:             for client in self.clients do
18:                 client.train-pred()
19:             end for
20:             self.load-global-model() for
21:             client in self.clients do
22:                 client.load-client-model()
23:             end for
24:         end for
25:     end for
26:     Plot accuracy, AUC, standard deviation of accuracy, and standard deviation of AUC
27: end procedure
```

**Algorithm 2.** The workflow of the cross-validation for the FedHome framework. The red lines are the contributions of this paper.

The experiments were conducted on a computer with an NVIDIA GTX 1060 GPU and an Intel Core i5 6500 3.6 GHz processor. Each sampling technique was tested over 200 training rounds, and the resulting performance metrics, such as test accuracy, AUC, and their respective standard deviations, were recorded and analyzed.

## V. RESULTS AND DISCUSSION

Based on the public implementation of this FedHome framework available in [18], we have tested different sampling techniques and evaluated the performance of each of them, applied to the HAR (Human Action Recognition) dataset. Then, in the next section, we will interpret these results.

Figure 4 shows the performance measures of the base model of the FedHome framework, before the personalization phase. The term AUC in these figures means Area Under the ROC Curve, quantifying a model's ability to distinguish between positive and negative instances in a binary classification task. It measures the overall "goodness" of the ROC curve, a graph depicting the trade-off between correctly classifying true positives and incorrectly classifying false positives. A higher AUC indicates a stronger separation between the two classes, meaning the model is more likely to correctly classify instances regardless of the chosen classification threshold.

Figures 5 to 10 shows the performance measures of the personalized model after applying different sampling techniques to the model. Finally, Figure 11 shows the standard deviation of Test Accuracy for different sampling methods in a violin graph.

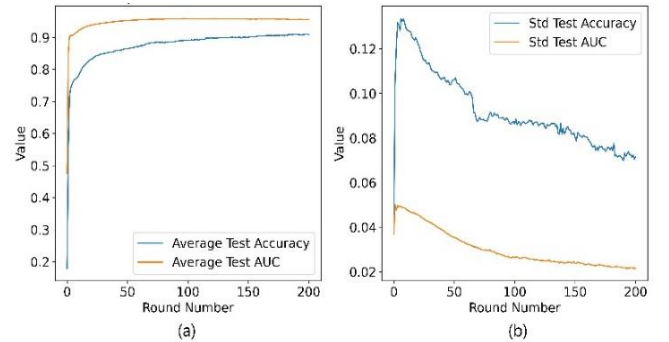

**Fig 4.** Illustration of how oversampling and undersampling techniques modifying and adding samples of minority class.

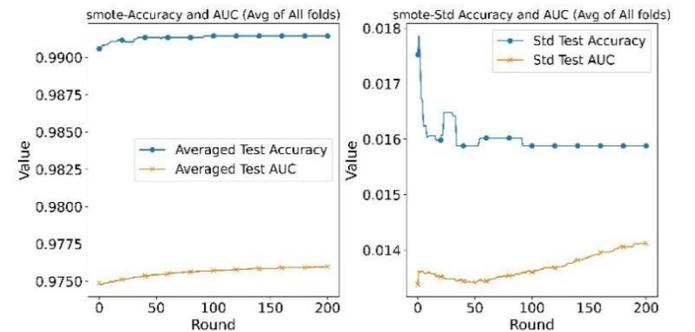

**Fig. 5.** Evaluation metrics for the personalized model after applying SMOTE as the oversampling technique.

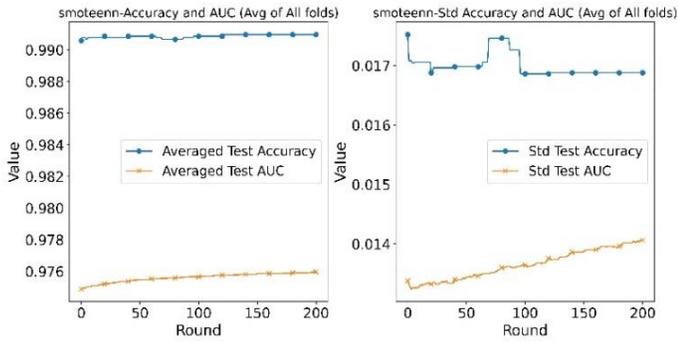

**Fig. 6**. Evaluation metrics for the personalized model after applying SMOTE-ENN as the oversampling technique.

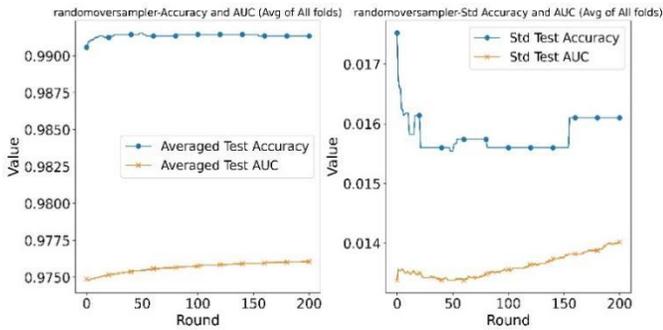

**Fig. 7**. Evaluation metrics for the personalized model after applying random oversampler as the oversampling technique.

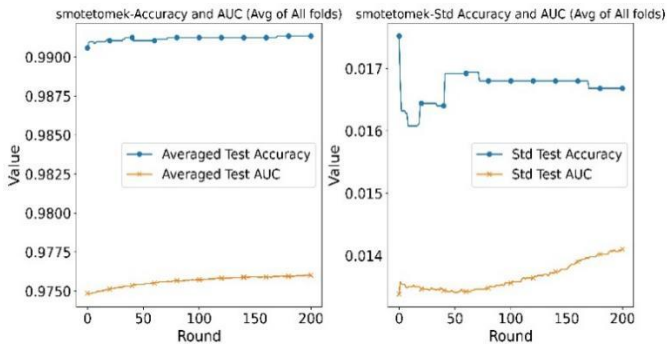

**Fig. 8.** Evaluation metrics for the personalized model after applying SMOTE-Tomek as the oversampling technique.

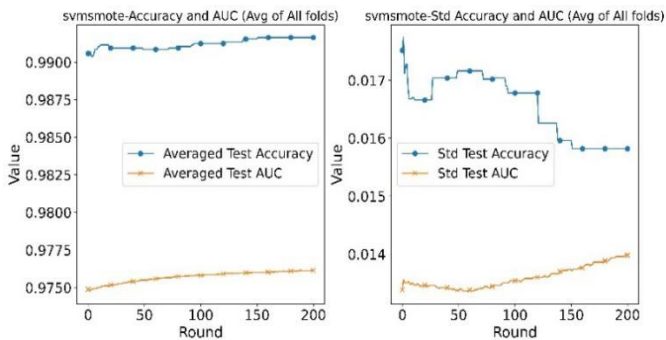

**Fig. 9.** Evaluation metrics for the personalized model after applying SVM-SMOTE as the oversampling technique.

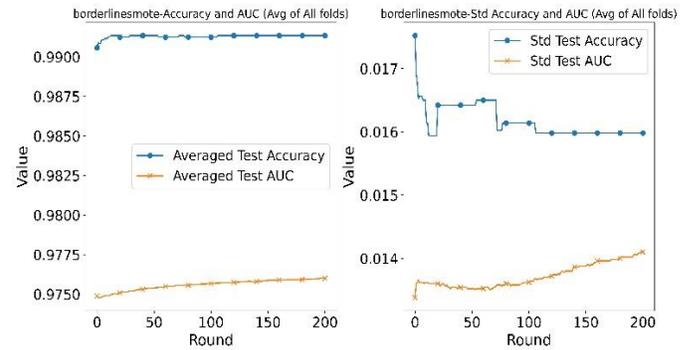

**Fig. 10.** Evaluation metrics for the personalized model after applying Borderline-SMOTE as the oversampling technique.

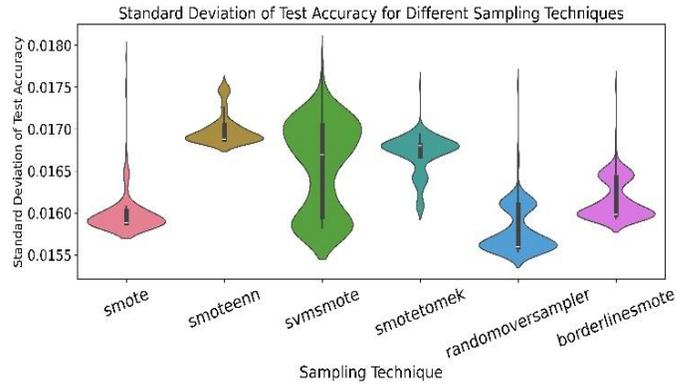

**Fig. 1** Average of standard deviation of test accuracy for different sampling methods from all folds.

In our analysis of various sampling methods within the FedHome framework, we found that the average accuracy of the personalized model consistently hovers around 98.8-99%. However, our study revealed notable variations in the standard deviation of test accuracy, a crucial metric for evaluating model performance, as highlighted in studies like [19].

SMOTE and SVM-SMOTE exhibited the highest standard deviation. Specifically, SMOTE ranged from 0.0157 to 0.0180, while SVM-SMOTE varied from 0.0155 to 0.0180. These fluctuations could stem from the inherent complexity of these methods. Both SMOTE and SVM-SMOTE involve oversampling the minority class, but SVM-SMOTE adds an additional layer of SVM-based cleaning. The interaction between oversampling and data cleaning likely contributes to higher variability in the datasets and, consequently, in the test accuracy.

Similarly, the Random Oversampler showed a notable standard deviation, ranging from 0.0155 to 0.0176. This method, which involves replicating instances in the minority class randomly, is prone to overfitting. As a result, the model may perform exceptionally well on some datasets that closely resemble the training set but falter on those that differ, leading to this heightened variability in test accuracy.

On the other hand, SMOTE-Tomek and SMOTE-ENN demonstrated more stable standard deviations. SMOTE-Tomek ranged from 0.0160 to 0.0175, while SMOTE-ENN ranged from 0.0167 to 0.0176 in test accuracy. SMOTE-Tomek combines SMOTE with Tomek link removal, and SMOTE-ENN combines SMOTE with Edited Nearest Neighbors (ENN) rules. ENN generally removes more examples than Tomek links, leading to a smoother decision boundary. This might be the reason behind SMOTE-ENN's consistent standard deviation in test accuracy.

## VI. Conclusion

In this study, we have evaluated and compared various sampling techniques with cross-validation to address class imbalance in personalized in-home health monitoring within the FedHome framework. FedHome, rooted in federated learning, employs a generative convolutional autoencoder to develop a global model from decentralized data collected by edge devices in residential settings. Our evaluation employed diverse metrics, including averaged test accuracy, Area Under Curve, and their respective standard deviations, to measure the efficacy of these sampling techniques.

Our analysis indicates that SMOTE-ENN exhibits a consistent standard deviation in test accuracy, translating to a more stable performance across different sampling scenarios. Such stability is invaluable in real-world applications where consistent performance is highly desirable. Conversely, SMOTE and SVM-SMOTE showed the highest standard deviation in test accuracy, highlighting greater variability and potential sensitivity to changes in data samples. This level of variability might pose challenges in contexts requiring robustness against data distribution shifts.

The Random Oversampler, with its high standard deviation, suggests a propensity for overfitting and limited generalizability. Overfitting, a common issue in machine learning, leads to models that excel with the training data but falter with new, unseen data. Therefore, the risk of overfitting is critical to consider when selecting a sampling method.

SMOTE-Tomek and SMOTE-ENN demonstrated more stable performance, with SMOTE-ENN having the lowest standard deviation among all methods evaluated. This consistent performance makes SMOTE-ENN a particularly attractive option for applications where reliability is crucial. SMOTE-Tomek, while slightly more variable, still showed better stability compared to Random Oversampler and other methods.

Considering our findings, we recommend SMOTE-ENN as the primary sampling technique for FedHome, as it provides a balanced approach between oversampling and undersampling, and creates a smoother decision boundary. However, choosing the best sampling technique may depend on specific application needs and dataset characteristics. Therefore, we suggest thoroughly analyzing data distributions and performance metrics before applying any particular sampling technique.

## Future Work and Open Problems

While this work compares sampling techniques for FedHome's personalized health monitoring, there are several promising directions for future work. Evaluating the communication efficiency of different sampling methods in federated settings (real-life scenarios) is an interesting direction since efficiency is a key concern in distributed learning. Investigating advanced generative models like GANs for synthetic minority sample generation may also unlock new capabilities. Another promising area is developing online sampling strategies that dynamically adapt sampling to incoming data.

Considering multi-objective sampling to optimize metrics like accuracy, efficiency, and diversity simultaneously could make the methods more holistic. Tailoring oversampling and undersampling techniques to FedHome's GCAE model architecture may lead to further improvements. Rigorously analyzing the privacy implications of oversampling techniques in federated learning is also an open research question.

Finally, expanding the analysis to other personalized monitoring applications beyond health, such as wearables, medical IoT, smart homes, etc., would help generalize the findings. This paper provides a starting point for understanding sampling in federated learning-based health monitoring systems. The analysis can be extended in numerous promising directions to enhance minority-class further learning under distributed privacy constraints.